\documentclass[11pt,a4paper]{article}

\PassOptionsToPackage{breaklinks}{hyperref}
\usepackage[draft]{hyperref}

\usepackage[hyperref]{acl2019}
\usepackage{times}
\usepackage{latexsym}
\usepackage{CJKutf8}
\usepackage{amsmath}
\usepackage{graphicx}
\usepackage{float}
\usepackage{multirow}
\usepackage{makecell}
\usepackage{url}
\aclfinalcopy 

\title{Modeling Intent, Dialog Policies and Response Adaptation for~Goal-Oriented~Interactions}

\author{ Saurav Sahay \quad Shachi H Kumar \quad Eda Okur \quad Haroon Syed \quad Lama Nachman \\
Intel Labs, Anticipatory Computing Lab, USA \\
\tt\small{\{saurav.sahay, shachi.h.kumar, eda.okur, haroon.m.syed, lama.nachman\}@intel.com}}

\begin{document}
\maketitle
\begin{abstract}
Building a machine learning driven spoken dialog system for goal-oriented interactions involves careful design of intents and data collection along with development of intent recognition models and dialog policy learning algorithms. The models should be robust enough to handle various user distractions during the interaction flow and should steer the user back into an engaging interaction for successful completion of the interaction. In this work, we have designed a goal-oriented interaction system where children can engage with agents for a series of interactions involving `Meet \& Greet' and `Simon Says' game play. We have explored various feature extractors and models for improved intent recognition and looked at leveraging previous user and system interactions in novel ways with attention models. We have also looked at dialog adaptation methods for entrained response selection. Our bootstrapped models from limited training data perform better than many baseline approaches we have looked at for intent recognition and dialog action prediction.

\end{abstract}

\section{Introduction}

Language technologies have benefited from recent progress in AI and Machine Learning. There have been major advancements in spoken-language understanding \cite{mikolov2013efficient,mesnil2015using}. Machine-learning approaches to dialog management have brought improved performance compared to traditional handcrafted approaches by enabling systems to learn optimal dialog strategies from data \cite{paek2008automating,bangalore2008learning}. With availability of large amounts of data and advancements in deep learning research, end-to-end trained systems have also shown to produce state of the art results in both open-ended \cite{dodge2015evaluating} and goal-oriented applications \cite{bordes2016learning} in the research community.

With the emergence of reliable ASR and TTS systems, we are now seeing platforms such as Alexa and Google Home and a plethora of domain and task-based dialog agents that allow users to take specific actions via spoken interface-based systems. Microsoft Research released Language Understanding Intelligent Service (LUIS) \cite{williams2015fast,williams2015rapidly}, which helps software developers create cloud-based, machine-learning powered, language-understanding models for specific application domains. Google Dialogflow is an SDS platform for quick development of various conversational agents that can be deployed on various platforms such as Amazon Alexa, Google Home and several others. Systems like Google's Dialogflow offer mechanisms such as explicit context management on linear and non-linear dialog flows to manage the conversations. The developer can attach input and output context states explicitly on various intents and create if-then-else flows via context variables. To go from explicit context management to implicit/automatic context management in SDS, probabilistic and neural network based systems are emerging in the research and development community~\cite{bocklisch2017rasa,burtsev-etal-2018-deeppavlov,ultes-etal-2017-pydial}.

Most dialog systems are categorized as either chatbots or task-oriented where chatbots are open-ended to allow generic conversations and the task-oriented system is designed for a particular task and set up to have short conversations \cite{Jurafsky:2000:SLP:555733}. Goal-oriented Interaction Systems are somewhat midway between the two and should support longer duration interactions having various tasks to fulfill as well as support some non-task interactions. 

We are developing a goal-oriented multimodal conversational system that engages 5 to 7 years old children in concise interaction modules \cite{DBLP:conf/icmi/AndersonPSMACRS18}. The overall goal of the system is to engage children in multimodal experiences for playful and learning oriented interactions. 

\begin{figure}[ht]	
	\centering
	\includegraphics[width=\linewidth]{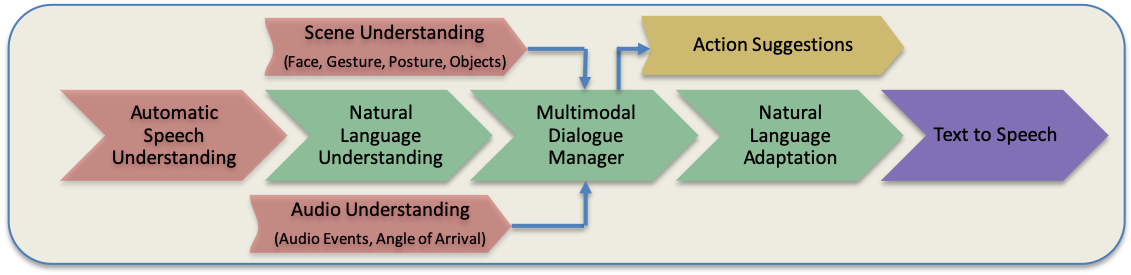}
	\caption{Multimodal Spoken Dialog System}
	\label{fig:app}
\end{figure}

Our application consists of interactions where children get introduced to the agent and they can play a simplified version of `Simon Says' game with the agent. The dialog manager ingests verbal and non-verbal communication via the Natural Language Understanding (NLU) component (entities and intents) and other engines that process vision and audio information (faces, pose, gesture, events and actions) and generates sequential actions for utterances and non-verbal events. We describe the NLU, Dialog Manager (DM) and {Dialog Adaptation} modules in this work as shown in Figure~\ref{fig:app}. We build our NLU and DM based models on top of the Rasa framework \cite{bocklisch2017rasa}. We enrich the NLU Intent Recognition model in Rasa by adding additional features to the model. Typical goal-oriented dialog managers are modeled as sequential decision making problems where optimal interaction policies are learned from a large number of user interactions via reinforcement learning approaches \cite{shah2016interactive, liu2017e2e, su-etal-2017-sample, levin1997stochastic, cuayahuitl2017simpleds, dhingra2016towards}. Building such systems to support children interactions is particularly difficult and we use a supervised learning approach using some initial training data collected via Amazon Mechanical Turk (AMT) to bootstrap the agents.  Dialog Adaptation has been explored as part of parametric generation \cite{mairesse2007personage} as well as end-to-end NLG for generating contextually appropriate responses \cite{duvsek2016context}. We look at parametric generation methods and develop a simple ML classifier for Dialog Adaptation.

\section{Data Collection}
\begin{figure}[!b]	
	\centering
	\includegraphics[scale=0.39]{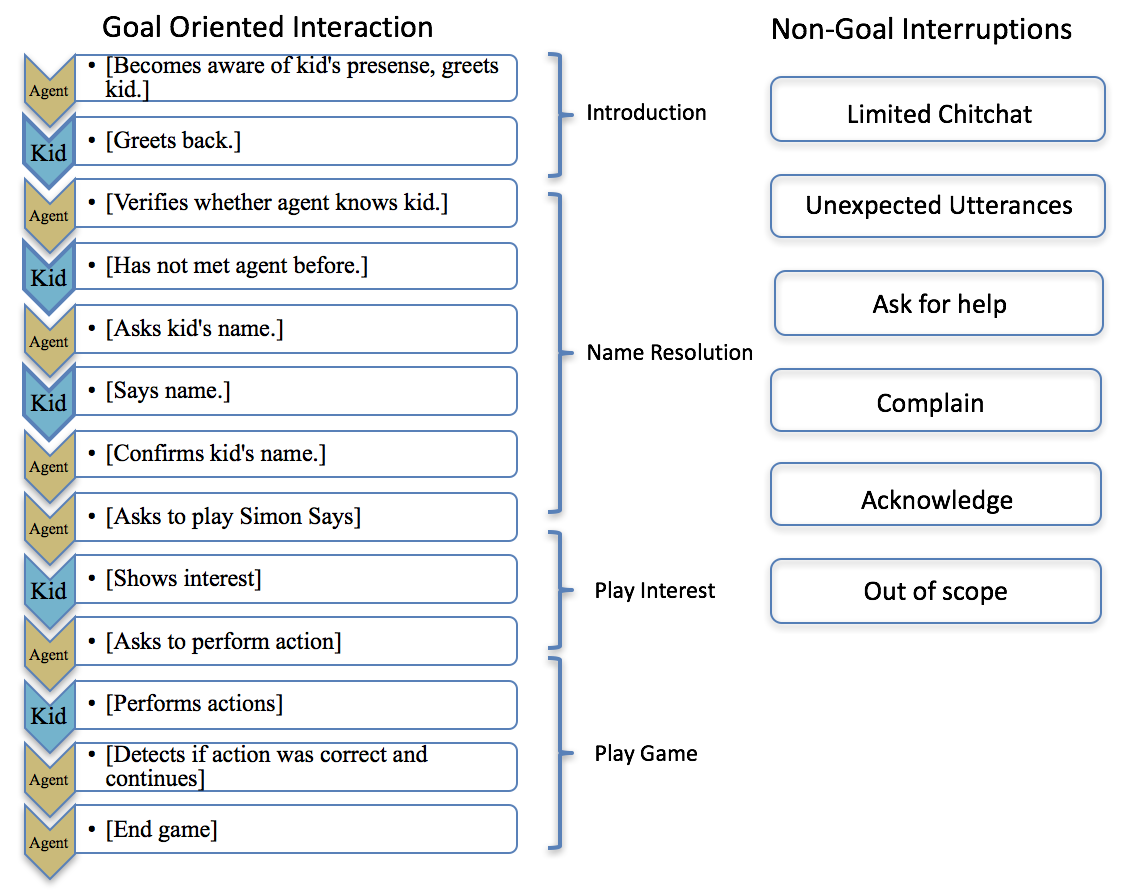}
	\caption{Domain of Application}
	\label{fig:dataCollect}
\end{figure}

Guided by User Experience (UX) research and Wizard of Oz user studies involving kids, we have designed sample interactions between the agent and the kid (Figure~\ref{fig:dataCollect}) for the  `Meet \& Greet' and `Simon Says' interactions.
The left half of Fig.~\ref{fig:dataCollect} shows the task-oriented dialog flow. However, in a real system, one could imagine a lot of non-task-oriented or non-cooperative dialog involved. Especially in a dialog involving young kids, there could be a lot of chit-chat style conversations on things such as preferences or likes/dislikes of kids, a description of how the day went at school, or even the kids asking several questions about the agent itself. As shown in the right half of the Fig.~\ref{fig:dataCollect}, these can be categorized as either simple chit-chat or some unexpected utterances, as well as complaints or requests for help. To support our application, which includes a mixture of task and non-task-oriented conversations, we collect data via AMT for two types of interactions: `Meet \& Greet', and `Simon Says', between the agent and the kid. We requested the turkers to provide us with a dialog, with agent on one side and kid on the other by providing sample interaction flows as shown in Fig.~\ref{fig:dataCollect}.

We collected 80 dialogs in total (65 stories for training, and 15 for test) for the two interactions. After the annotation process, we observed 48 user intents (including verbal intents and physical activities for the `Simon Says' game), as well as 48 distinct agent actions. Note that for our NLU models, 26 distinct verbal intents are observed in the dataset, where 16 of them are categorized as goal-oriented (i.e., related to our `Meet \& Greet' and `Simon Says' scenarios), and the remaining 10 are non-goal-oriented (e.g., chit-chat or out-of-scope). Further details of our NLU and dialog datasets can be found in Table~\ref{table:nlu-stats} and Table~\ref{table:dialog-stats}, respectively.

\begin{table}[t]
\scriptsize
\begin{center}
  \begin{tabular}{| l | c | c | }
    \hline
     & Training & Test \\
     \hline
     \# of intents (distinct) & 26 & 26 \\
     \multicolumn{1}{|r|}{goal-oriented} & 16 & 16 \\
     \multicolumn{1}{|r|}{non-goal-oriented} & 10 & 10 \\
     \hline
     total \# of samples (utterances) & 779 & 288 \\
     total \# of tokens & 3850 & 1465 \\
     \hline
     mean \# of samples per intent & 29.96 & 11.08 \\
     mean \# of tokens per sample & 4.94 & 5.09 \\
     \hline
  \end{tabular}
    \caption{NLU Dataset Statistics}    
    \label{table:nlu-stats}
\end{center}
\end{table}

\addtolength{\tabcolsep}{-1pt}
\begin{table}[t]
\scriptsize
\begin{center}
  \begin{tabular}{| l | c | c | c | c | c | c |}
    \hline
     & \multicolumn{3}{c|}{Training} & \multicolumn{3}{c|}{Test}\\
     \hline
     \# of dialogs & \multicolumn{3}{c|}{65} & \multicolumn{3}{c|}{15} \\
     \multicolumn{1}{|r|}{Meet \& Greet} & \multicolumn{3}{c|}{49} & \multicolumn{3}{c|}{12} \\
     \multicolumn{1}{|r|}{Simon Says} & \multicolumn{3}{c|}{16} & \multicolumn{3}{c|}{3} \\
     \hline
     & Kid & Agent & All & Kid & Agent & All\\
     \hline
     \# of intents/actions & 48 & 48 & 96 & 28 & 28 & 56\\
     \multicolumn{1}{|r|}{goal-oriented} & 39 & 41 & 80 & 19 & 22 & 41\\
     \multicolumn{1}{|r|}{non-goal-oriented} & 9 & 7 & 16 & 9 & 6 & 15 \\
     \hline
     \# of turns & 441 & 560 & 1001 & 97 & 112 & 209\\
     \multicolumn{1}{|r|}{goal-oriented} & 374 & 501 & 875 & 63 & 84 & 147\\
     \multicolumn{1}{|r|}{non-goal-oriented} & 67 & 59 & 126 & 34 & 28 & 62 \\
     \hline
     \# of turns per dialog & 6.8 & 8.6 & 15.4 & 6.5 & 7.5 & 14.0\\
     \hline
  \end{tabular}
    \caption{Dialog Dataset Statistics}    
    \label{table:dialog-stats}
\end{center}
\end{table}
\addtolength{\tabcolsep}{+1pt}

\section{Models and Architecture} \label{models}
In this section, we describe the architectures we developed for the NLU, DM and Dialog Adaptation modules of the spoken dialog system pipeline. 
\subsection{NLU/Intent Understanding} \label{nlu-models}
The NLU module processes the input utterance to determine the user intents and entities of interest. 

\subsubsection{Intent Classifier}
We use an intent classifier based on supervised embeddings provided as part of the Rasa framework \cite{bocklisch2017rasa}. This embedding-based intent classifier is based on ideas from the StarSpace algorithm \cite{wu2017starspace} and embeds user utterances and intents into the same vector space. These embeddings are trained by maximizing the similarity between them. We also adapt sequence models and Transformer networks\footnote{https://github.com/RasaHQ/rasa/pull/4098} to work with various features for developing our models.   

\subsubsection{Features and models for the NLU module}
We utilize and extend the Rasa NLU module by adding various textual features and models to improve the performance of the intent classifier. 


\textbf{Textual features:} We used text features such as number of words, first word, last word, bigrams, dependencies such as 1st/2nd/3rd person subject, inverted subject-verb order and imperative verbs, morphology features, hand constructed word lists, `wh' words, top n words, and many more. We add about 580 such textual features to our custom feature extractor.

\textbf{Speech Act features:} Verbal Response Modes (VRM) is a principled taxonomy of speech acts that can be used to classify literal and pragmatic meaning within utterances \cite{lampert2006classifying}. Utterances are classified into disjoint sets comprising Question, Disclosure, Edification, Advisement, Acknowledgement, Reflection, Interpretation and Confirmation according to this model\footnote{For the classifier, Disclosure and Edification classes were combined into one class}. The classifier \cite{sahay2011intentional} also used the above text features for modeling the task and the top features in this classification task were domain independent features such as `?', length of words, `you', `i',`okay', `well', etc. 
      
\textbf{Universal Sentence Embeddings}:
Universal Sentence Encoders  \cite{DBLP:conf/emnlp/CerYKHLJCGYTSK18} encode sentences into high dimensional vectors that has shown success in a variety of NLP tasks. We use the encoder model trained using a Transformer encoder\footnote{https://tfhub.dev/google/universal-sentence-encoder-large/3} to generate fixed length vectors as features for the NLU module in our pipeline. The motivation for using this model is to hope to recognize short utterances with similar meaning where the word level vectors may not provide enough information for correct intent classification. 

\textbf{Sequence Models}:
Long Short Term Memory (LSTM) networks \cite{DBLP:journals/neco/HochreiterS97} and Bidirectional LSTMs \cite{bi-lstm-1997} can capture patterns and dependencies in sequential data using their memory gates and can robustly encode information. We use LSTM and BiLSTM layers to generate representations that are used in place of the fully connected embedding layer in the baseline model.

\textbf{Transformers}:
\cite{DBLP:conf/nips/VaswaniSPUJGKP17} proposed a novel sequence-to-sequence  network, the Transformer,
entirely based on attention mechanism. The performance of Transformer model has generally surpassed RNN-based models and achieved better results in various NLP tasks. We use rich representations from the transformer model as an extension to the baseline model.

\textbf{Bidirectional Encoder Representations from Transformers (BERT)}:
Bidirectional Encoder Representations from Transformers, BERT \cite{devlin2018bert} represents one of the latest developments in pre-trained language representation and has shown strong performance in several NLP tasks. We use pre-trained BERT model based features to generate representations for our dataset and use this in our NLU pipeline.


\subsection{Dialog State Tracking}
\begin{figure}[ht]	
	\centering
	\includegraphics[width=\linewidth]{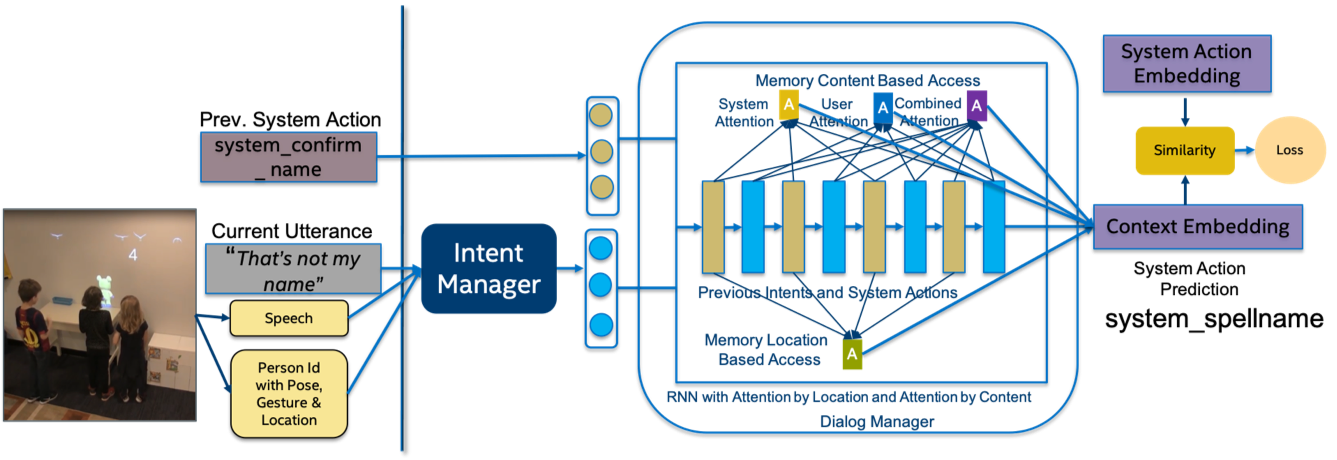}
	\caption{Dialog State Tracking Architecture}
	\label{fig:dst_arch}
\end{figure}

The task of a Dialog Manager is to take the current state of the dialog context and decide the next action for the dialog agent by using some policy of execution. Policies that are learnt from actual conversational data either use Markov Decision Processes (MDP) or Memory augmented Neural Networks. Memory augmented networks can update the context of the dialog by storing and retrieving the dialog state information. The dialog states and past system actions can be encoded as domain knowledge in the network to encode dialog manager based rules using Hybrid Code Networks (HCN) \cite{DBLP:journals/corr/WilliamsAZ17}. One such dialog management policy that combines memory operations with domain knowledge embedding is the Recurrent Embedding Domain Policy (REDP) \cite{vlasov2018few}. This policy represents the dialog state embedding as a recurrent network with explicit management of the user memory and system memory modules. The user memory is used to attend to previous user intents and the system memory is used to attend to previous system actions. REDP uses Bahdanau Attention scoring to attend to previous user utterances (user memory) and memory address by location as developed in Neural Turing Machines (NTM) for system memory. With NTM, the previous system actions can be copied directly into the final dialog state representations to help recover from a sequence of uncooperative user utterances. REDP learns embeddings for dialog and system actions in the same space using ideas from the StarSpace algorithm \cite{wu2017starspace} and uses these embeddings to rank the similarity between the recurrent embedding and the system action. Figure \ref{fig:dst_arch} shows the overall architecture of the dialog state tracker.

Our application handles multimodal streams of high frequency non-verbal input such as person recognition via face and speaker identification, gestures, pose and audio events. We pass on the information via separate modules to the dialog manager (bypassing the NLU) as relevant intents for goal-oriented interaction.  

\begin{figure}[ht]	
	\centering
	\includegraphics[width=\linewidth]{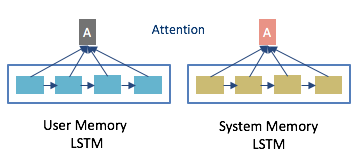}
	\caption{User Memory and System Memory}
	\label{fig:app1}
\end{figure}


\subsubsection{User and System Memories}
Attention-based models \cite{DBLP:journals/corr/BahdanauCB14} can dynamically retrieve relevant pieces of information via selective reading through a relatively simple matching operation. In the REDP baseline architecture, separate mechanisms are used for attending to past user utterances and past system actions. While the system memory block helps the agent recover from uncooperative dialog by attending to and copying past system actions (using NTMs), the user memory mechanism uses Bahdanau Attention based matching to attend to relevant parts of past user utterances and enhance the user intent signal in the LSTM.

\subsubsection{Fusion of User and System Memory}
In this work, we capture the previous user-system interactions into the recurrent architecture by fusion of the signals from the user inputs and system actions. We hypothesize that attending to previous combinations of user utterances and system actions can help the bot choose the right action by directly leveraging multiple discrete views of the dialog context information. This may be useful in contexts involving multi-party interactions. Agents can also benefit from discrete attention in situations where deviations from the task-oriented dialog flow can lead to small multi-turn interaction where the context of the dialog (combinations of previous user utterances and responses) is crucial.

\begin{figure}[ht]	
	\centering
	\includegraphics[width=\linewidth]{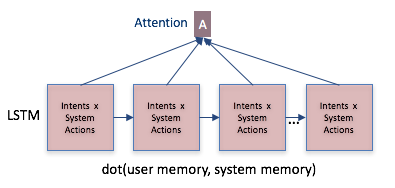}
	\caption{Capturing interactions between intents and system actions}
	\label{fig:app2}
\end{figure}

Figure~\ref{fig:app1} shows the memory units based on previous user intents and system actions. Figure~\ref{fig:app2} shows the fused memory unit obtained using the dot product of user memory and system memory. It computes the product of all possible interactions between the intent features and system action features to obtain a memory of size [\emph{user embedding x system action embedding}]. 
This results in an enhanced representation of the dialog embedding vector (\ref{TDot} below). This dialog embedding vector is further enhanced using NTM based copying of the relevant previous system action as described in \cite{vlasov2018few}. 

We incorporate another fusion technique as described in \cite{zadeh2017tensor} where one larger memory component explicitly represents user memory, system memory and the interactions between the two (\ref{TFN} below).

We create the following combinations of the user and system memory blocks as part of the LSTM cell for attending to these memories for computing the relevant system action:

\begin{enumerate}
\itemsep-0.2em 
\item Concatenation of User and System memories (Single Memory Unit)
\item \label{TDot} Tensor Dot of User and System memories (Single Memory Unit)
\item  \label{TFN} Tensor Fusion of User and System memories (Single Memory Unit)
\item Separate Attention on User and System memories (Two Memory Units)
\item Separate Attention on User memory and Tensor Dot (Two Memory Units)
\item \label{goodConfig}Separate Attention on User memory, System memory and Tensor Dot (Three Memory Units)
\item Separate Attention on User memory, System memory and Tensor Fusion (Three Memory Units)
\end{enumerate}


\begin{figure}[ht]	
	\centering
	\includegraphics[width=\linewidth]{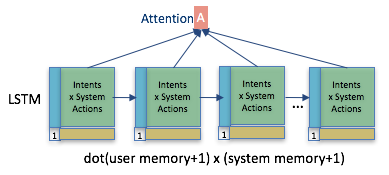}
	\caption{User Memory x System Memory Fusion}
	\label{fig:app3}
\end{figure}

\begin{figure}[ht]	
	\centering
	\includegraphics[width=\linewidth]{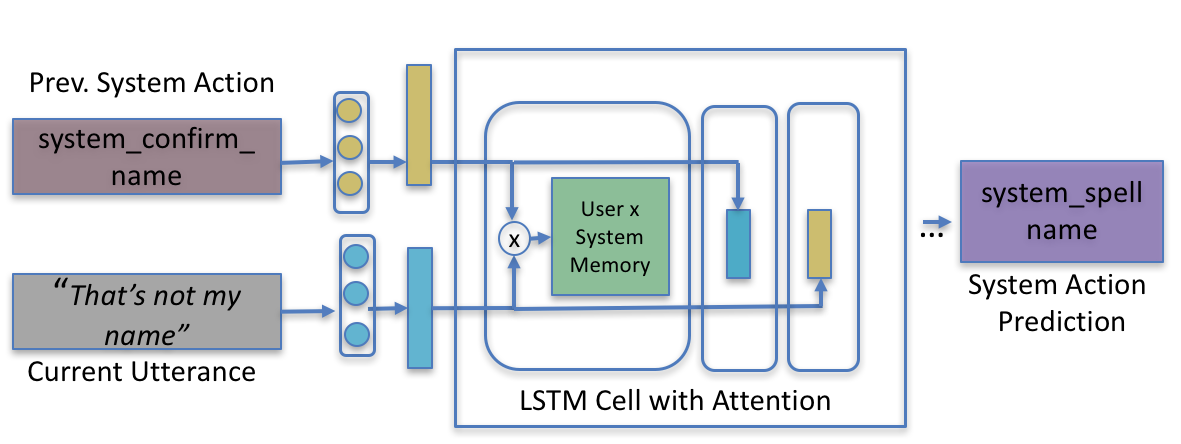}
	\caption{Simplified View of Contextual Attention}
	\label{fig:app4}
\end{figure}

 The configurations in the list above are a combination of the user memory, system memory and the fused user-system memory blocks. Configuration \ref{goodConfig} above is also conceptually shown in Figure \ref{fig:app4} with separate attention blocks to user memory, system memory and the combined user and system memory blocks.

\subsection{Response Adaptation}
\begin{figure}[ht]	
	\centering
	\includegraphics[width=\linewidth]{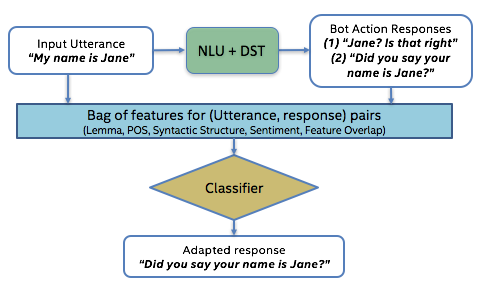}
	\caption{Response Adaptation}
	\label{fig:SDS}
\end{figure}

Children adapt to syntax and words used by their dialog partner more frequently than adults \cite{10.1007/978-3-642-15760-8_67}. This phenomenon is called Entrainment and generally applies to copying the conversational partner's attributes related to phonetics \cite{doi:10.1121/1.2178720}, syntax \cite{Reitter06primingof}, linguistic style \cite{doi:10.1177/026192702237953}, postures, facial expressions \cite{article}, etc. It has also been linked to overall task success \cite{176df8b190e843ccb56d78a61b3997b0} and social factors \cite{doi:10.1177/0956797610392928}. In this work, we explore lexical and syntactic adaptation, by using the similar referring expressions or sentence structure. SDS could pick a response at random from a list of responses for actions to match the predicted dialog state. In our work, we use the response adaptation module to score the responses and choose the best response instead of picking any random response. Figure \ref{fig:SDS} shows our architecture for response adaptation. 

\section{Experiments and Results}
In this section we present and discuss the results for the NLU and DST experiments based on the architectural explorations explained in Section \ref{models}. 
\subsection{NLU Experiments}
To evaluate the feature explorations for the NLU model, we present an ablation study on the LSTM, BiLSTM, Transformer, USE, text + speech acts features (SA) and BERT feature additions presented in Section \ref{nlu-models}. We use the embedding intent classifier as the baseline for the NLU experiments. For sequence models and Transformer model, we use word level features instead of the sentence-level features. Table~\ref{table:nluAblation} shows the precision, recall and F1-scores for the overall intent identification using different features. 

The performance of the baseline model with the text + speech act features added is very similar to the baseline. Adding USE to the baseline improves the scores and this can be explained from the fact that USE, which is a transformer based model, generates better representations compared to individual text/speech act features. Consistent with the fact that BERT has shown a lot of success in several NLP tasks recently, we notice that adding BERT representations alone to the baseline improves the performance significantly. On the word-level features, we observe that while LSTM, BiLSTM and Transformer representations show slight improvement, adding BERT along with these shows a significant performance increase. In general, from Table~\ref{table:nluAblation}, we can conclude that the speech act features have a very minor impact on the performance. The results show that adding BERT with the Transformer model helps achieving best performance on the dataset.

Table~\ref{table:nlu_qualitative} presents some qualitative comparison of the baseline model with the text + speech acts (B+SA), USE (B+USE) and BERT (B+BERT) features against the ground truth labels (GT). From the table, we see that a short utterance such as `i am okay' is labeled as `mynameis' by the B+SA model. We believe that this can be attributed to the text features that look at words/phrases such as `i am', and the `mynameis' intent would usually start with `i am'. B+USE model predicts the correct label for this utterance. Although B+BERT model assigns the incorrect label, the prediction is semantically close to the GT. 
We again observe that a very short phrase such as `oh shit' is classified incorrectly by both the B+SA and B+BERT models. We believe that for very short utterances, the USE model generates better representations as compared to BERT, and hence can produce more meaningful predictions. 

An interesting point to observe is that, for utterances such as `that ain't me Oscar' or `are you Alexa?', the BERT feature based model associates them with the `deny' and `useraskname' intents, respectively. Although these intents are wrongly identified, they are semantically very close to the ground truth labels `wrongname' and `askaboutbot'. 
While the B+SA model tends to generate incorrect predictions for challenging examples, the B+USE model classifies `are you Alexa?' as out-of-scope. Longer utterances such as `where is the nearest shopping mall' are well handled by the BERT model, while the other models fail. We can conclude that the USE model could better handle very short sentences, and the BERT model performs better on the longer ones.

\renewcommand{\arraystretch}{1.2}
\addtolength{\tabcolsep}{+1pt}
\begin{table}[t]
\small
\begin{center}
\resizebox{\columnwidth}{!}{
  \begin{tabular}{| l | c  c  c | }
    \hline
     & Prec  & Rec &  F1  \\ 
     \hline
     Baseline (StarSpace)  & 0.74 &  0.65 &  0.66 \\ 
     \hline
      Baseline + LSTM  & 0.75 &  0.70 &   0.70         \\ 
      Baseline + BiLSTM  & 0.76 &  0.70 &   0.70         \\
      Baseline + Transformer  & 0.72 &  0.70 &   0.68         \\ 
      Baseline + BERT + BiLSTM  & 0.84 &  0.81 &   0.81         \\ 
      \textbf{Baseline + BERT + Transformer}  & \textbf{0.87} &  \textbf{0.82} &   \textbf{0.83}         \\ 
 \hline
     Baseline + SA & 0.74 & 0.68 & 0.68 \\ 
     Baseline + USE  & 0.76 & 0.71 & 0.72 \\ 
     Baseline + BERT & 0.81 & 0.77 & 0.77 \\
     \hline
     Baseline + USE + SA  & 0.76  & 0.72 & 0.73 \\
     Baseline + USE + BERT & 0.82 & 0.76 & 0.77 \\
     \hline
     Baseline + USE + SA + BERT  & 0.83 & 0.78 & 0.78\\
     \hline
  \end{tabular}
  }
    \caption{NLU Ablation Experiments on Meet \& Greet and Simon Says Dataset}   
    \label{table:nluAblation}
\end{center}
\end{table}
\addtolength{\tabcolsep}{-1pt}
\renewcommand{\arraystretch}{1}

\begin{table*}
\small
\begin{center}
  \begin{tabular}{| l | c | c | c | c |}
    \hline
   Kid's utterance & B+SA & B+USE  & B+BERT  &  GT\\  \hline
  `i am okay'  & my-name-is &  user-doing-good &   ask-how-doing  &     user-doing-good     \\ \hline
 
     `oh shit' & ask-how-doing & user-missed-it &  affirm  &     user-missed-it     \\ \hline

      `are you Alexa?' & ask-how-doing &  out-of-scope  &   user-ask-name &  ask-about-bot    \\ \hline
 
       `that ain't me Oscar' & out-of-scope  &  my-name-is  &  deny  &     wrong-name    \\ \hline

  
       `where is the nearest shopping mall' & next-step &  user-i-am-sick  &  out-of-scope  &     out-of-scope    \\ \hline

  \end{tabular}
    \caption{Qualitative analysis of baseline with USE+SA as features vs baseline with BERT}   
    \label{table:nlu_qualitative}
\end{center}
\end{table*}

\subsection{Dialog State Tracking Experiments}

\begin{figure}[!b]
	\centering
	\includegraphics[scale=0.3]{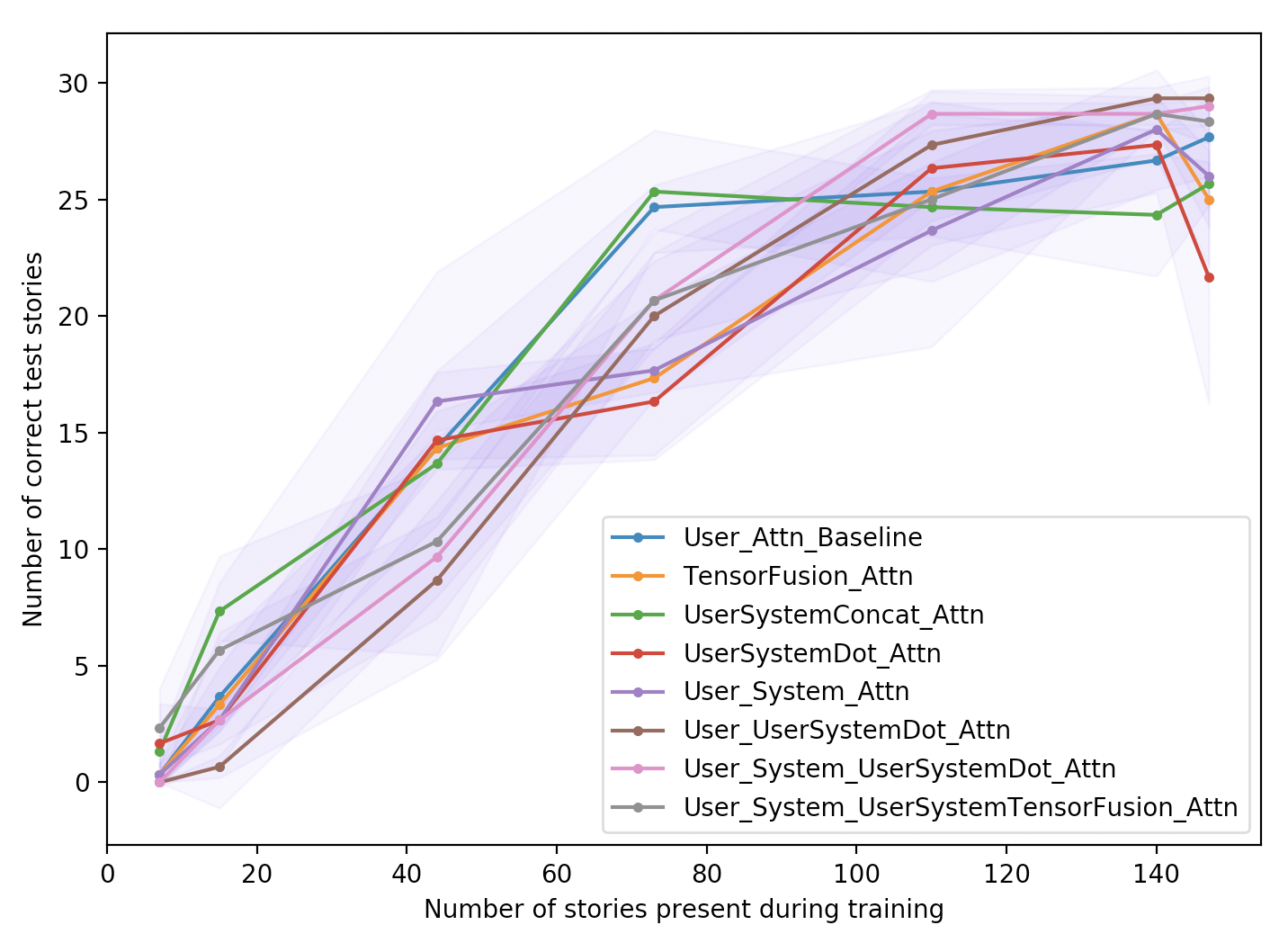}
	\caption{Performance of Models with RNN Attention over User and System Memory configurations with varying Training Sizes}
	\label{fig:neurips}
\end{figure}

We investigate the role of single vs. multiple memories for attention as well as the impact of system memory and user memory fusion with the policy explorations. In Figure \ref{fig:neurips}, we compare the results of the baseline REDP policy from \cite{vlasov2018few} with our policy changes on the dataset used by the authors that contain uncooperative and cooperative dialogs from hotel and restaurant domain. We use the same test set from the hotel domain and use a combination of cooperative and uncooperative dialogs from both hotel and restaurant domain for the training set. We divide the training set into 7 splits with 0, 5, 25, 50, 70, 90, 95 percent exclusion in the number of dialog stories in the domains. The baseline policy from \cite{vlasov2018few} applies Bahdanau Attention scoring to the history of user utterances only. The policy does not explore attending to previous system actions or combinations of those for updating the RNN states as part of Bahdanau Attention scoring. In our experiments, we reuse the NTM based memory copying mechanism for system actions but explore additional effects of leveraging previous system actions and their combinations with the previous user intents. We see that using separate attention blocks on user memory, system memory and the combined memory using their dot product interactions help achieve slightly improved performance on this dataset. We can see some advantages in the case of very limited training data (when the agent cannot perhaps copy previous system actions) as well as in the case of full training set, where we see a slightly higher number of correct test stories with other policies. Further investigation is needed to understand if the proposed policy changes in REDP would always benefit in certain scenarios. We also try to investigate the comparison between using a single (larger) memory attention vs. using multiple memory attentions. For example, as shown in Figure \ref{fig:neurips}, 3 policy changes perform better than the baseline policy, all of which use separate memory blocks and all of these attend to the fused interaction representation of user and system memory.

\begin{figure}[!b]	
	\centering
	\includegraphics[scale=0.5]{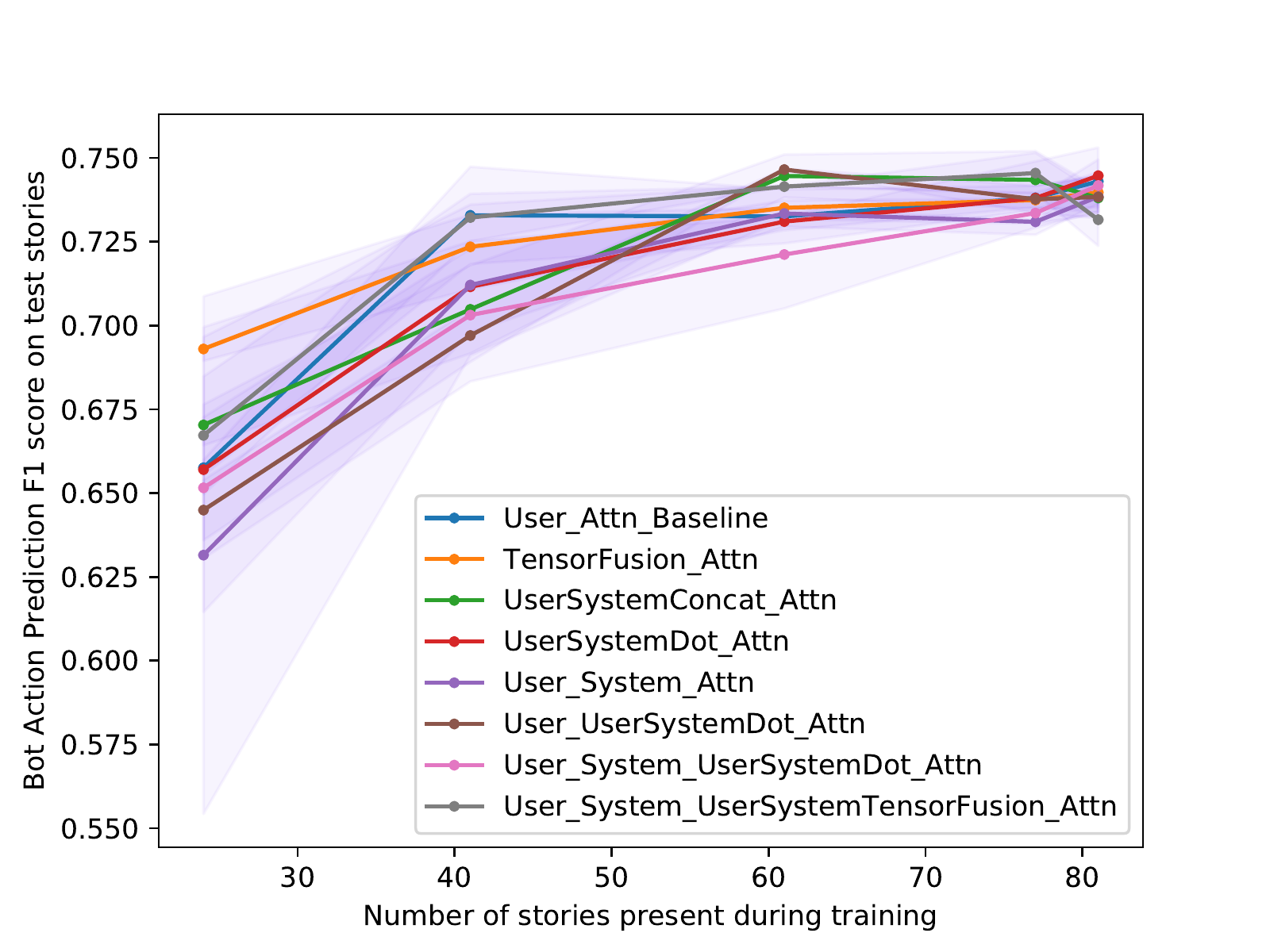}
	\caption{Action Prediction F1-score of Models with RNN Attention over User and System Memory configurations with varying Training Sizes}
	\label{fig:ksdialog}
\end{figure}

Figure \ref{fig:ksdialog} shows the action prediction F1-scores for our `Meet \& Greet and Simon Says' dataset. Our test stories were collected using AMT and we allowed the bot and user to engage in any number of utterances and actions before passing control to the other. Since we also did not also impose any particular sequence in the game play, we didn't expect to get all action predictions correct in any one of test stories. We show the F1-scores for action predictions for the test stories varying the size of the training set. The more training data the agent has seen, more user utterance and system action interactions it has seen capturing application regularities, therefore we can hope to see improved bot performance on unseen test sets with multiple memories and fusion configurations of attention units. From Figure \ref{fig:ksdialog}, we can only say that there is a lot of variance in predictions with lesser training data, general trend for all these policies is get better with more training data. We see that the overall best F1-score is achieved with 75\% of the training data with two separate attentions, one for user memory and another for the user-system fusion interactions. 

\subsection{Dialog Adaptation}
For generating contextually appropriate responses, we collected multiple responses for bot utterances and trained ML models for selecting the most appropriate response from the list of responses. The goal of the classifier was to match the syntactic and linguistic style of the speaker. For this, we used 60 dialogs, with 32400 instances (2753 positive instances, 29647 negative instances) and 243 features. We created positive and negative instances automatically using feature overlap counts between the context dialog and the responses to be adapted. For feature generation, we extracted lemmas, Parts of Speech, Syntactic Structures and Sentiment Polarities using Stanford CoreNLP suite \cite{manning-EtAl:2014:P14-5}.  

\begin{figure}[ht]	
	\centering
	\includegraphics[width=\linewidth]{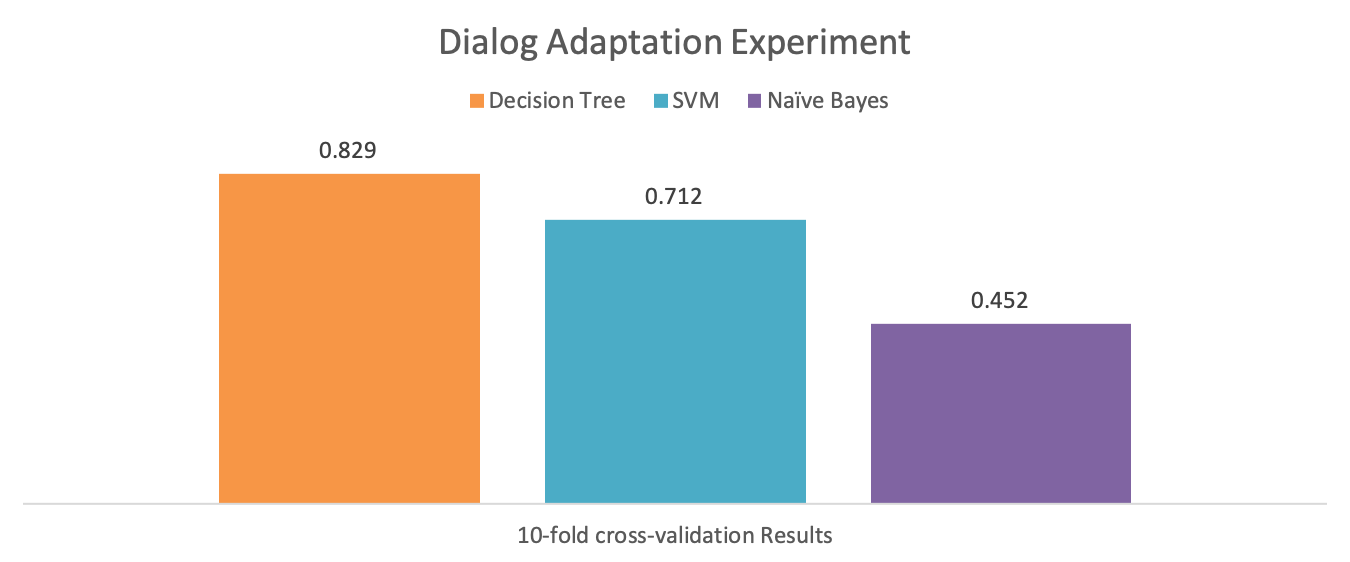}
	\caption{Dialog Adaptation}
	\label{fig:dadapt}
\end{figure}

\begin{figure}[ht]	
	\centering
	\includegraphics[width=\linewidth]{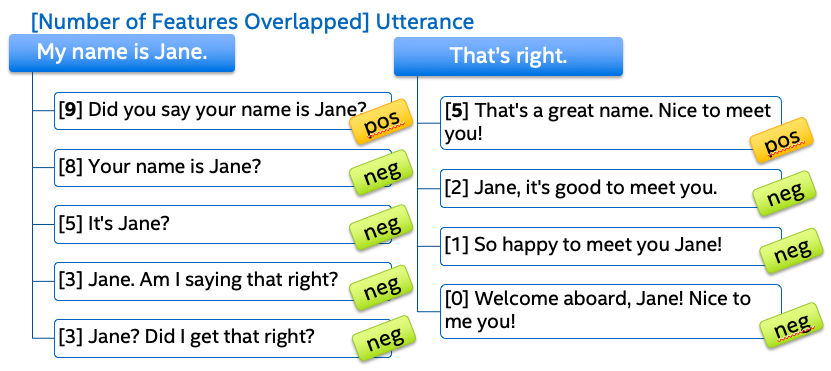}
	\caption{Dialog Adaptation Examples}
	\label{fig:dadaptexample}
\end{figure}

Figure \ref{fig:dadapt} shows the cross-validation performance of the classifiers on the dataset. We use the Decision Tree based classifier for response adaptation in the spoken dialog system. Figure \ref{fig:dadaptexample} shows a couple of response adaptation examples along with the positive and negative classification results for the two context utterances.

\section{Conclusions \& Future Work}
We report preliminary explorations and results for our data driven spoken dialog system development for the multimodal `Meet \& Greet and Simon Says' goal-oriented application. The application involves phases of interactions for introduction, name resolution, game related interaction and actual game play involving children. We collect NLU and Dialog Data for our application using AMT, and manually identify non-goal-oriented intents and design interactions to include various non-goal-oriented or `uncooperative' paths in the  interaction. We have used and extended the Rasa NLU module and Rasa Core module for Dialog Management.
Our application involves five to seven year-old children communicating with agents and we have seen from data that many children use very short utterances. In order to have a robust NLU, we have explored the use of lexical, syntactic and speech act related features (SA features), Universal Sentence Encoders as well as BERT embeddings for the embedding-based intent classifier which is a  part of the Rasa NLU stack.
We see the largest improvement in the NLU performance using the pre-trained BERT features and the Transformer model. For Dialog State Tracking, we extended the REDP policy by including different configurations of User and System Memory for RNN based Attention. We looked at a method for Single Memory Unit Tensor Fusion for combining User Memory, System Memory and tensor fused representation of User and System Memory. We explore other multiple memory unit configurations for RNN based Attention on history of User Intents, System Actions and their combinations. We saw improvements over the REDP baseline policy for the hotel and restaurant domain dataset as well as the `Meet \& Greet and Simon Says' dataset. We also explored Response Selection from the list of response templates as an Adaptation Classification problem using features such as Lemma/POS, Syntactic feature overlap and Sentiment of the response. As part of future work, we plan to extend the NLU and DM based models to include multimodal information in the pipeline. 

\section*{Acknowledgments}
Special thanks to Glen Anderson and the Anticipatory Computing Lab Kid Space team for conceptualization and the UX design for all the interactions. We thank Zhichao Hu (UCSC) who worked as a Summer Intern with us in 2017 and worked on the Dialog Adaptation module. We greatfully acknowledge and thank the Rasa Team and community developers for the framework and contributions that enabled us to further our research and build newer models for the application. 

\bibliography{acl2019}
\bibliographystyle{acl_natbib}

\end{document}